# Evaluating AI Meeting Summaries with a Reusable Cross-Domain Pipeline

Philip Zhong, Don Wang, Jason Zhang

Webex Suite AI
Cisco Systems, Inc.
San Jose, California, USA

{lizhon, dongdwan, xiaojzha}@cisco.com

*The views expressed in this paper are those of the authors and do not necessarily reflect the views of Cisco.*

## Abstract

Industrial teams often deploy large language model features before stable regression or model selection evaluation exists. We present a reusable evaluation system for AI meeting summaries that combines structured ground-truth (GT) construction, fixed candidate generation, claim-grounded scoring, persisted reporting, and a privacy-bounded online monitoring and nomination interface. The online evidence is not itself a benchmark: privacy-safe aggregate exports show active monitoring, hard regime detection, and directional movement without exposing customer data. We benchmark the offline path on `114` meetings across `city_council`, `private_data`, and `whitehouse_press_briefings`, yielding `340` completed meeting-model pairs and `680` judge runs for `gpt-4.1-mini`, `gpt-5-mini`, and `gpt-5.1`. Under this fixed protocol, accuracy differences are not statistically significant under Holm correction (corrected p-values `0.053-0.448`), although `gpt-4.1-mini` has the highest mean accuracy (`0.583`); the significant separation is on retention, where `gpt-5.1` leads on completeness (`0.886`) and coverage (`0.942`). Typed slices isolate `whitehouse_press_briefings` as an accuracy-hard regime, and a later focused rerun over `gpt-4.1`, `gpt-5-mini`, and `gpt-5.4` reuses the same stack under the same judges and metrics. This extended preprint keeps those core results aligned with the formal submission while adding a more detailed repository-level account of cross-domain reuse from the companion AI-search paper and an additional typed `DeepEval` contrastive analysis.

**Model naming note.** Running text uses canonical model names on first introduction. Tables, filenames, and artifact IDs retain compact report labels for consistency with the packaged benchmark outputs. Table A maps the two conventions and is repeated in Section 4.3 where candidate-generation settings are defined.

Table A. Compact report labels used in tables, filenames, and packaged artifact IDs.

| Compact report label | Canonical model name | Primary use in this paper |
|---|---|---|
| `gpt-41-mini` | `gpt-4.1-mini` | Main merged benchmark tables and artifact rows |
| `gpt-5-mini` | `gpt-5-mini-2025-08-07` | Main merged benchmark tables and artifact rows |
| `gpt-51` | `gpt-5.1` | Main merged benchmark tables and artifact rows |
| `gpt-41` | `gpt-4.1` | Focused deployment follow-up |
| `gpt-54` | `gpt-5.4` | Focused deployment follow-up |

## 1. Introduction

Meeting-summary features are easy to demo and hard to govern. Once a product team starts swapping foundation models, tuning prompts, or reacting to user complaints, informal spot checks stop being enough. The operational questions become concrete: which model should be the default, which failures are benchmark problems rather than model problems, and how can production feedback be turned into reusable regression tests instead of one-off triage?

Meeting corpora and meeting-summary benchmarks such as ICSI, AMI, QMSum, MeetingBank, and ExplainMeetSum already establish the task context, while factuality work shows why unsupported details must be surfaced explicitly rather than hidden behind a single holistic score. Against that backdrop, the practical gap is not another benchmark alone but a deployment-facing governance workflow that remains stable across model swaps and release reviews.

This paper describes the evaluation system used for AI meeting summaries in Webex Suite AI, a large-scale deployed enterprise product surface. The system is designed as a reusable quality loop rather than as a standalone scorer. It persists transcripts, GT artifacts, candidate summaries, judge outputs, and comparison reports; it evaluates models under a fixed protocol; and it supports the maintenance path from online incidents back into offline benchmark refresh. At that scale, even modest regressions in default summary quality can have broad user-visible and business-visible impact, so model-switch decisions need more than spot checks.

The same orchestration substrate has also been reused in adjacent evaluation tasks. In AI-search evaluation, the control-plane pattern supported fixed-protocol comparison of embedding choice, chunking, reranking, and indexing on a synthesized benchmark integrated with CI/CD. That earlier system paper, *Evaluating Embedding Models and Pipeline Optimization for AI Search Quality* by Philip Zhong, Kent Chen, and Don Wang (arXiv:2511.22240), established the repository-level evaluation pattern for a retrieval-centered task. The present paper instantiates the same control plane for AI meeting summaries, where the reference artifact becomes structured meeting-summary GT and the comparison layer becomes claim-grounded factual scoring over generated summaries.

This continuity is repository-level rather than merely conceptual. Across the search and meeting-summary papers, the stable components are persisted artifacts, controlled batch execution, report-backed comparison, judge/result aggregation, significance reporting, and release-facing evaluation outputs. What changes across domains is the semantic layer rather than the control plane: search uses query-document relevance references and retrieval metrics, whereas meeting summarization swaps in structured GT generation, application-facing summary candidates, and claim-grounded factual evaluators. We therefore present the meeting-summary benchmark not as an unrelated application study, but as a second domain instantiation of the same evaluation substrate under a changed task-specific reference and scoring layer.

The positioning is system-oriented. Relative to meeting-summary benchmark papers, this work contributes an operational loop rather than a static dataset. Relative to factuality-metric papers, it contributes a reusable, release-facing workflow for deployed meeting-summary quality governance rather than a standalone scorer. Relative to general-purpose evaluation frameworks, it treats typed GT construction, paired significance reports, a persisted benchmark-artifact layer, and an explicit online-to-offline nomination interface as first-class parts of the system rather than surrounding utilities.

This extended preprint is written as a **system paper with companion detail**. Its main empirical claims are aligned with the formal submission, but it elaborates the repository-level reuse story more explicitly and keeps a typed `DeepEval` contrastive baseline as additional context. The central contribution is not a new factuality metric in isolation, but a reusable evaluation substrate implemented in the underlying Dataset Pipeline system and surfaced here through persisted artifacts, benchmark reports, and selected analysis scripts.

Relative to the earlier AI-search paper, the present work contributes three new system elements. First, it elevates **structured GT generation** into an explicit pipeline stage rather than assuming static reference data. Second, it introduces **typed benchmark partitioning**, which makes it possible to surface domain-specific failure regimes rather than only merged averages. Third, it adds **claim-level structured scoring with formal metric definitions**, which in turn enables the significance-backed model-selection analysis and the White House error-structure diagnosis later in the paper. These additions are not cosmetic extensions of the earlier pipeline; they are the mechanisms that make the meeting-summary benchmark diagnostically useful.

The main contributions are:

1. a five-stage reusable evaluation control loop (source intake → structured reference construction → candidate generation → structured scoring → persisted reporting) that enables cross-domain reuse by separating orchestration from task-specific reference schemas, demonstrated through the meeting-summary instantiation in this paper and the companion AI-search instantiation
2. an implemented meeting-summary evaluation stack, exposed in this public package through transcript-backed artifacts, structured GT outputs, structured score outputs, benchmark reports, and selected analysis/export scripts
3. an expanded typed benchmark over `114` meetings and three candidate models, including dataset-type breakdowns, meeting-level significance testing, and evaluator disagreement analysis
4. a later focused deployment rerun over `city_council` and `private_data` showing that the same evaluation system can be reused, without changing judges or metrics, to answer a narrower model-selection question over `gpt-4.1`, `gpt-5-mini`, and `gpt-5.4`

# 2. Design Goals and Cross-Domain Reuse

The design is driven by four system goals.

## 2.1 G1: Inspectable Artifacts

Every major object in the pipeline should be auditable after execution: source transcript, GT, candidate summary, judge outputs, and report. This favors file-backed persistence over ephemeral evaluation runs.

## 2.2 G2: Reusable Orchestration, Task-Specific References

The orchestration layer should be reusable across AI tasks even when the reference artifact changes. In AI search, the reference is query-document relevance. In meeting summarization, the reference is structured GT over meeting topics, points, and decisions. The pipeline should keep the orchestration stable while swapping the task-specific reference layer and metric definitions.

## 2.3 G3: Release-Oriented Comparison

The pipeline should support repeated comparison of candidate systems under a controlled protocol, so that model-selection decisions are backed by persisted evidence rather than one-off manual review.

## 2.4 G4: Extensibility from Offline to Online Quality Loops

The quantitatively validated core in this paper is the offline loop. The current architecture also includes an implemented online path, comprising feedback ingestion, evaluation triggering, result persistence, and downstream metrics emission, in which production issues can be clustered, curated, and turned into new benchmark items. That path is not evaluated here as a controlled online benchmark, but it is supported by retained aggregate operational evidence. In the deployed Webex Suite AI environment, this path is driven by user feedback signals and production quality escalations rather than by synthetic replay alone. In this sense, the system is designed not only for evaluation but for evaluation maintenance.

A compressed view of the retained online aggregates illustrates the intended maintenance role. After deduplicating a 30-day transcript-quality export by the latest record per meeting ID, `2,178` meetings remain; `75.0%` score at least `4` on overall transcript quality, while only `23.1%` do so on fluency. Applying the retained hard-regime rule (`OverallQuality <= 2` and `NoiseRatio >= 30`) nominates `254` cases for offline refresh, averaging `1.58` overall quality, `1.94` fluency, and `2.35` comprehensibility. A separate sanitized two-month aggregate shows notes thumbs-up moving from roughly `69%` to `81%`, action-item thumbs-up from roughly `78%` to `82%`, and structure/no-details issues falling from the low `20s` to single digits. These signals are aggregate operational evidence for monitoring, pattern discovery, and candidate nomination; they do not carry the main offline benchmark conclusions.

These goals yield a clean cross-domain abstraction. The reusable skeleton is:

- persist source artifacts rather than treating them as transient inputs
- construct task-specific reference artifacts under version control
- generate or ingest candidate outputs under controlled configurations
- evaluate candidates with structured, inspectable outputs
- persist reports for model selection, regression tracking, and future dataset improvement

Two properties are especially important in the present meeting-summary instantiation.

- **structured GT generation**: the repository does not treat GT as an opaque annotation blob; it generates and persists typed meeting objects
- **structured score generation**: the evaluator does not emit only a single free-form judgment; it emits claim-level alignments, per-metric scores, explanations, and issue labels that can be aggregated later

This design also aligns with the internal roadmap concepts behind the broader AI datasets and evaluation program:

- **golden seed data**: high-quality reference artifacts anchor trustworthy benchmarking
- **dataset synthesis and quality assurance**: data generation is paired with validation rather than treated as a blind scaling step
- **evaluation orchestration**: repeated runs are controlled by a common execution pattern
- **evaluation metrics and insights**: the system must produce not only scores, but diagnosis and actionable model-selection signals

## 2.5 Positioning Against General-Purpose Evaluation Frameworks

General-purpose evaluation frameworks such as RAGAS [10, 11], TruLens [12], and DeepEval [13] solve adjacent problems and are valuable points of comparison. RAGAS emphasizes reusable metrics and synthetic testset generation. TruLens emphasizes feedback functions, tracing, and ground-truth agreement. DeepEval emphasizes metric libraries, local test execution, and CI/CD-oriented LLM testing. The present system overlaps with all three, but its scope is different: it treats structured GT construction, structured score generation, typed benchmark partitioning, and persisted release-facing artifacts as first-class pipeline stages rather than optional surrounding infrastructure.

The comparison in Table 0 is architectural rather than exhaustive. It is intended to clarify system scope, not to claim that other frameworks cannot be extended. Capabilities evolve quickly; the table summarizes the official literature where available and otherwise the official software documentation cited in the references as accessed on April 17, 2026. We do not claim an empirical head-to-head evaluation against these systems in the present paper; that comparison is deferred to future work. Here, `Partial` indicates that a framework offers a functionally adjacent capability but not the specific combination of structured generation, typed partitioning, and file-backed persistence that this system treats as first-class. These ratings are based on the official literature where available and otherwise the official documentation cited in References 11-13, and they may not reflect later framework updates.

Table 0. Architectural capability comparison between the present system and adjacent general-purpose evaluation frameworks. The purpose is to clarify systems scope, not to claim score superiority.

| Capability dimension | This system | RAGAS | TruLens | DeepEval | Standalone claim scorer |
|---|---|---|---|---|---|
| File-backed end-to-end artifacts | Yes | No | Partial | Partial | No |
| Reusable cross-domain architecture | Yes | Partial | Partial | Partial | No |
| Structured GT generation stage | Yes | Partial | No | No | No |
| Meeting-summary-specific typed schema and metrics | Yes | No | No | Partial | Partial |
| Typed dataset partitions | Yes | No | No | No | No |
| Release-gating / regression reports | Yes | No | No | Partial | No |
| Multi-model comparison reports | Yes | Partial | Partial | Partial | No |
| CI/CD quality-gate path | Yes | No | No | Partial | No |

# 3. System Architecture

The repository is not itself a meeting assistant product. It is the evaluation and artifact layer that supports one.

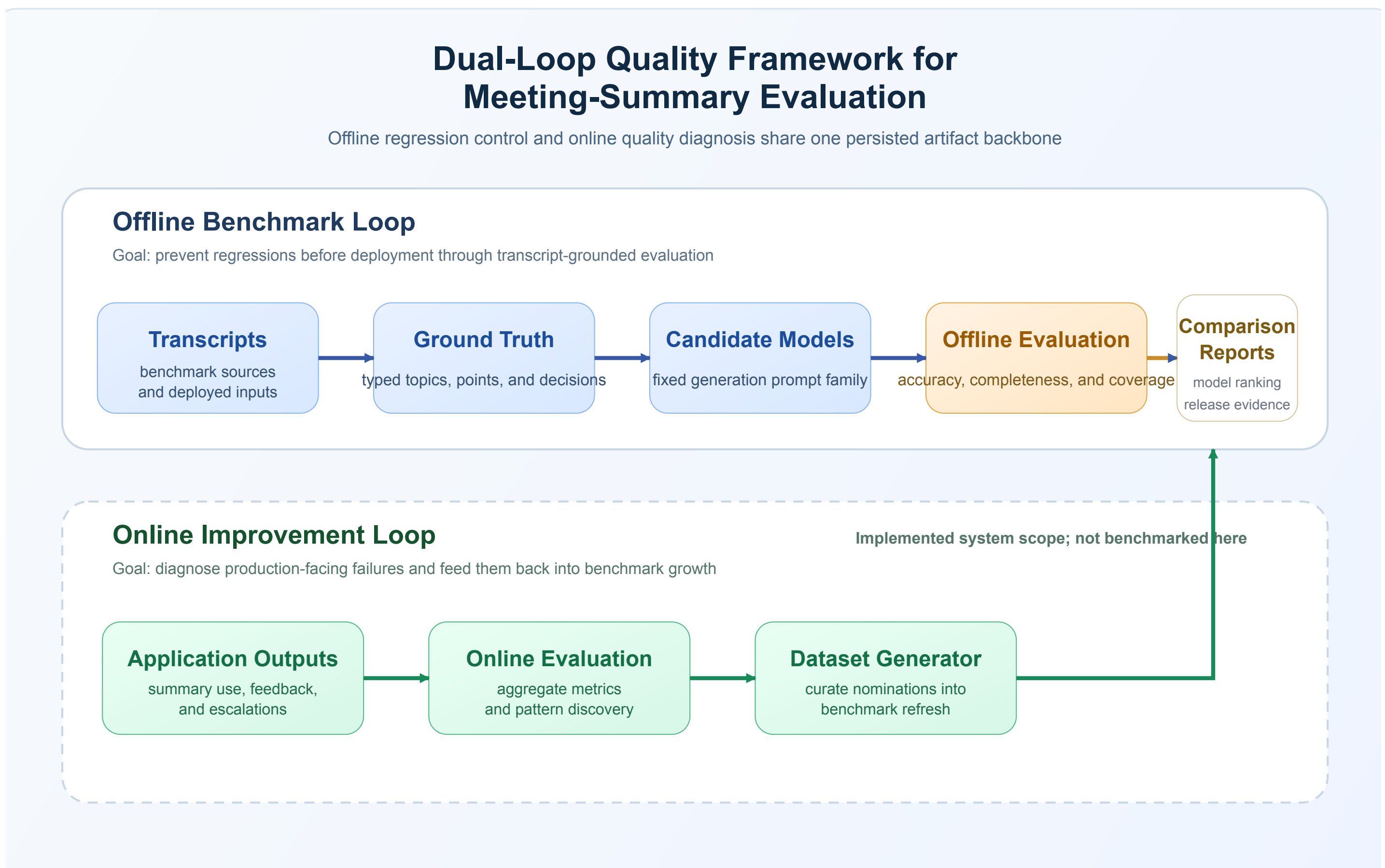

Figure 1. Reusable quality-loop architecture for AI meeting-summary evaluation. Solid components correspond to the benchmarked offline loop. The dashed online loop is implemented as system context but is not benchmarked in this paper.

Figure 1 shows the high-level architecture. Read through the roadmap lens, the loop consists of five reusable stages: a transcript simulator or production-data intake layer, a dataset generator, an offline evaluation stage, a CI/CD quality gate, and an online evaluation stage. The offline loop provides benchmark-backed regression control and is the portion validated directly in this paper. The online loop is implemented in the current system as an integrated feedback-to-evaluation path, but it is not quantitatively benchmarked in this paper. In the deployed Webex Suite AI environment, the online path is triggered by user feedback signals and production quality escalations; curated cases are periodically promoted to new benchmark items to refresh the offline regression suite. We therefore include it as deployed system context while restricting the paper's measured claims to the offline loop.

At the full-system level, this translates into transcript-backed artifact storage, prompt-versioned GT construction, prompt-versioned candidate generation, offline evaluation with structured judge outputs, entry points for online and grounded evaluation, and file-backed reports that remain inspectable after execution. The public package in this repository preserves the artifact/report layer and selected analysis scripts from that system. Taken together, these components support the three concrete benefits emphasized in the roadmap: benchmarking under a stable protocol, regression detection when models or prompts change, and failure-case discovery through persisted explanations and issue labels.

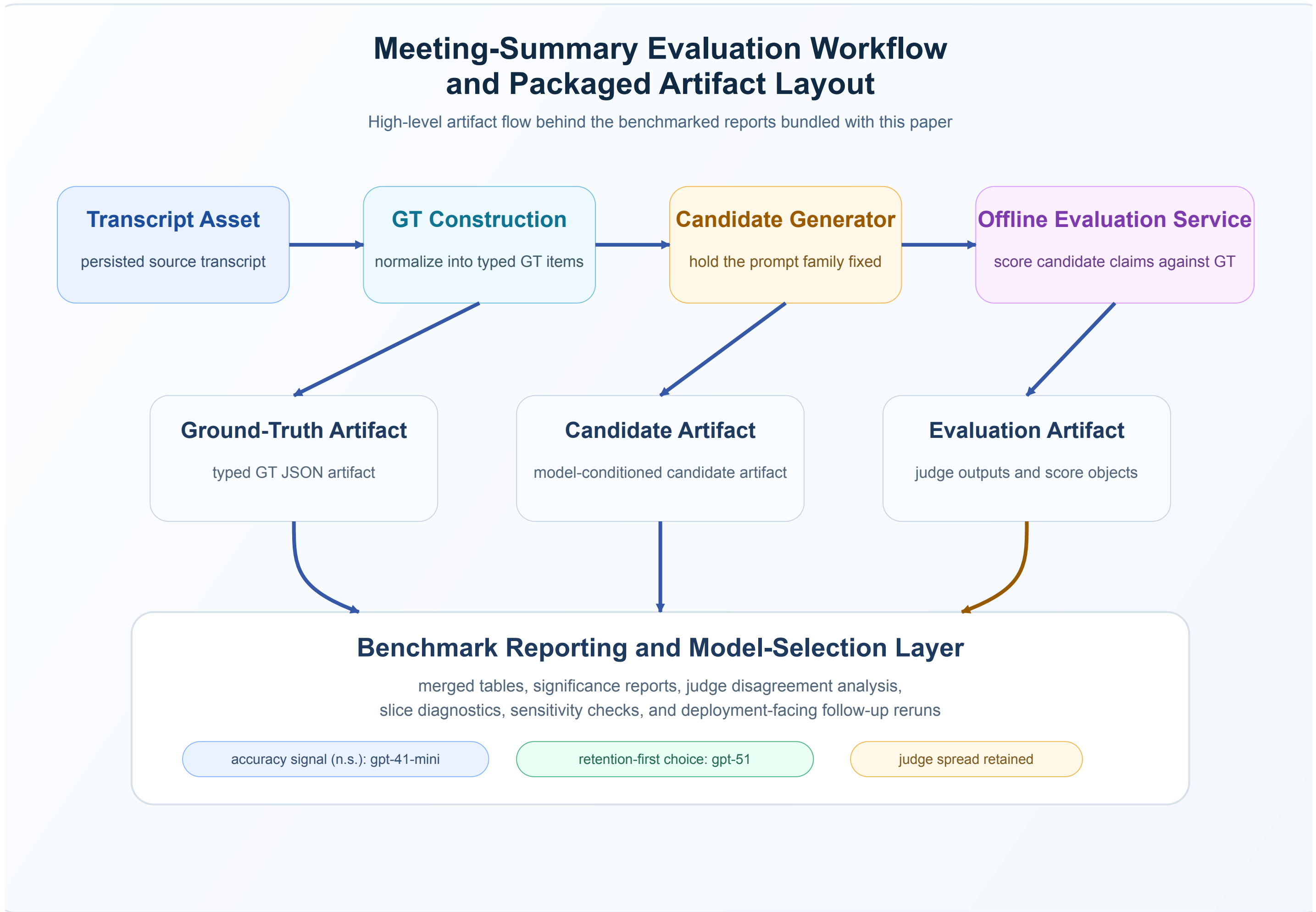

Figure 2. Workflow schematic for the meeting-summary evaluation pipeline and its packaged artifacts. Transcript assets feed both GT construction and candidate generation; offline evaluation consumes (GT, candidate) pairs and emits structured reports for comparison and deployment-facing analysis. The public package exposes the persisted artifact/report layer, while some upstream service modules remain internal.

Figure 2 grounds the design in the current artifact package. Unlike Figure 1, which is architectural, this view maps directly to the persisted artifacts and packaged scripts used in the benchmark workflow.

The two figures should be read together. In Figure 1, Stage 1 source intake corresponds to the transcript assets, Stage 2 structured reference construction corresponds to GT generation, Stage 3 candidate generation corresponds to application-facing summary generation, Stage 4 structured scoring corresponds to offline evaluation, and Stage 5 persisted reporting corresponds to the benchmark and release-report layer. In Figure 2, the transcript curation node takes raw transcript files as input and outputs normalized transcript assets. The GT construction node takes those transcript assets and outputs typed JSON structured around `meeting_context`, `participants`, `topics`, `points`, and `decisions`. The candidate-generation node takes the same transcript assets and outputs application-mirroring markdown summaries. The offline-evaluation node takes `(GT artifact, candidate artifact)` pairs and outputs structured evaluation JSON containing extracted claims, per-metric scores, issue labels, and explanations. This topology is deliberate: transcripts feed both GT and candidates so that reference construction and candidate generation remain comparable, while reporting stays downstream of structured scoring so that the same aggregation logic can be reused across domains.

# 4. Pipeline Components

## 4.1 Artifact Model

The packaged public artifacts in this repository use the following file-backed layout:

```
dataset/
├── assets/transcripts/internal/<meeting_id>/original_transcript.txt
└── views/meeting_notes/
    ├── ground_truth/internal/<meeting_id>/meetingsummary/ground_truth.json
    ├── candidate/internal/<meeting_id>/<variant>/<model>/baseline.md
    └── evaluation/internal/<meeting_id>/offline/evaluation_report_<variant>_<model>.json
```

This layout matters systemically because it makes every stage independently inspectable. Transcript, GT, candidate, and evaluator outputs can all be audited or regenerated without ambiguity. It also makes cross-domain reuse practical: the repository can keep the same orchestration while swapping task-specific reference artifacts. In roadmap terms, this is what allows the same quality loop to be aligned by different teams while preserving a common evaluation substrate.

In the packaged artifact tree, `dataset_type` is carried by the benchmark CSVs rather than by a directory-name level, and the raw `private_data` artifacts are intentionally omitted for confidentiality. As a result, the raw `dataset/` subtree in this public repository contains the `92` public meetings (`34` `city_council` + `58` `whitehouse_press_briefings`), while the aggregate benchmark CSVs still retain all `114` meetings, including the private slice. The same boundary applies to the later focused three-model rerun bundle: the released follow-up package preserves only derived aggregate CSV / Markdown reports and does not redistribute raw `private_data` transcript, GT, candidate, or evaluation files.

File-backed persistence is also a deliberate operational choice rather than a convenience format. Persisted files support independent inspection, partial regeneration, and version-controlled diff without requiring a live database connection months after execution. That property matters for release-facing evaluation pipelines because historical runs often need to be re-audited after the original execution context has disappeared.

The stage boundaries are similarly stable by design. The reference-construction stage consumes normalized transcript text and emits typed GT JSON; candidate generation consumes the same transcript asset and emits application-facing markdown; structured scoring consumes `(GT JSON, candidate markdown)` pairs and emits evaluation JSON containing metric scores, explanations, and issue labels. Prompt internals, model choices, and subset-specific heuristics are therefore implementation details behind persisted interface contracts rather than part of the orchestration boundary itself.

In this paper, **candidate** has a specific operational meaning: it is the application-generated meeting summary being evaluated against GT. In the current benchmark, candidates are produced by a repository pipeline that mirrors product behavior rather than by replaying production logs directly, but analytically they still represent the application's summary output.

## 4.2 Ground-Truth Construction

The meeting-summary GT is organized around:

- `meeting_context`
- `participants`
- `topics`
- `points`
- `decisions`

Point identifiers such as `t_001_p_001` are assigned after the semantic structure is generated. This preserves a stable downstream schema while keeping the generation stage focused on meeting content rather than identifier formatting.

At a high level, the GT path in this paper is a two-tier quality process: automated structured extraction produces the initial reference artifact, and bounded human review or automated audit strengthens that artifact when stronger evaluation trust is required. This framing matters because downstream model scores are only as reliable as the reference they are scored against.

This is a core systems property of the paper: GT is generated as a **structured representation**, not merely as untyped prose. That structured representation is what later enables stable claim alignment, metric aggregation, and typed benchmark analysis.

In the full implementation underlying this artifact package, the default GT path is automated. `GroundTruthService` calls `MeetingSummaryGroundTruthGenerator.generate_ground_truth_advanced()`, which executes a multi-stage pipeline rather than a single-pass extraction. Stage 1 generates two independent draft GTs, one using `gpt-5-2025-08-07` and one using `anthropic.claude-sonnet-4-20250514-v1:0`. Stage 2 aligns these drafts. Stage 3 re-reviews both uncertain and single-aligned items. Stage 4 merges the aligned and reviewed material, after which the repository assigns stable `t_###_p_###` and `d_###` identifiers and recomputes metadata counters. The exact prompt/config assets used for this benchmarked path are part of the internal implementation; the public package preserves the resulting artifacts and benchmark reports rather than the full prompt tree.

The dual-model drafting step is a systems decision, not just a prompt flourish. Single-model GT generation risks systematic omission of items that one model is less likely to extract, and those omissions would become silent reference loss. Dual-model drafting with explicit alignment turns disagreement into a first-class quality signal that can be reviewed, merged, or audited before it propagates into downstream scoring.

All stages are executed through JSON-schema-constrained prompts and cached as intermediate artifacts. Quality control in the default path is therefore primarily machine-based: schema-constrained generation, dual-model draft comparison, targeted review of uncertain items, identifier normalization, metadata recomputation, and an optional automated-audit path all act as guards on GT consistency. The repository also exposes an optional audit stage that scores GT faithfulness and coverage against the transcript, but that audit is not a mandatory gate for every item in the present benchmark. One subset is stronger than this default path: all 22 `private_data` GT artifacts in the merged typed benchmark underwent human review after automated construction. Those stronger `private_data` GT artifacts belong to the internal benchmark workspace and are not redistributed as raw files in the public `dataset/` subtree. The benchmark should therefore be interpreted as using machine-generated structured GT overall, with partial automated-audit retention across slices and an additional human-review layer for the full `private_data` subset.

The same automated GT generation pipeline is used across `city_council`, `private_data`, and `whitehouse_press_briefings`; there are no subset-specific model swaps in the reported benchmark. The main subset differences arise after generation. `private_data` receives an additional human-review pass over all benchmark items, whereas `city_council` and `whitehouse_press_briefings` remain on the automated path with optional automated audit only. In that stronger path, human review should be understood as **rubric hardening**, not as fully manual re-authoring from scratch. Reviewers inspect the same transcript against the structured GT draft, check for missing decisions, broken identifiers or schema structure, weak context, and topic-boundary errors, and then apply targeted corrections to move the artifact toward golden GT suitable for downstream scoring. The goal is not to replace automated GT generation with manual curation everywhere, but to make automated GT accurate enough that human review remains bounded. In a small internal spot-check, we also reviewed `11` anonymized private-slice GT-audit findings from five meetings spanning redundancy, topic placement, vague wording, a flagged omission, over-specific wording, and action-item overlabeling. Human review agreed with the automated audit in `10` cases outright and treated the remaining flagged omission as a lower-severity implementation nuance rather than a required summary point. Several of those agreeing cases were action-item downgrades where the automated audit correctly distinguished owner-bound follow-up work from milestone commitments, policy reminders, scheduling moves, timeline updates, or generic meeting reminders. We use that spot-check only as qualitative process validation rather than as a formal agreement study. A second difference is artifact normalization. `city_council` and `private_data` GT files persist metadata counters directly, whereas all `whitehouse_press_briefings` GT files retain the same structured topics and points but omit stored metadata counters, so evaluation traverses topics, points, and decisions at runtime. These distinctions matter for interpretation and are revisited in Sections 6.2 and 6.7.

Two transparency notes matter for interpretation. First, the exported report gt_agreement_stats_all_20260421.csv enumerates all `114` benchmark meetings, but only the `34` `city_council` rows retain populated stage-2 alignment counts. The `private_data` rows are present as coverage markers with `temp_dir_present_but_alignment_missing`, and the `whitehouse_press_briefings` rows are present as `no_retained_temp_artifacts`. Averaged over the populated `city_council` rows, those stage-2 reports record `15.91` fully aligned items, `13.53` single-aligned items, and `3.76` items sent to explicit review, while the final merged GT averages `32.79` output items. We report these as separate stage summaries rather than as a strict additive partition of final GT items, because review and merge can combine, drop, or re-express material before the final artifact is written. The follow-up review stages are conservative rather than cosmetic: reviewed single-aligned items are almost always kept or modified (`11.35` on average versus only `0.18` dropped), while reviewed uncertain items are usually merged (`3.03`) rather than excluded (`0.59`). These retained alignment reports provide a quantitative proxy for GT difficulty on the `city_council` slice. Benchmark-wide agreement percentages still cannot be reconstructed from the public package, because the non-`city_council` slices do not retain the same populated stage-2 counts. The `whitehouse_press_briefings` slice is therefore still the least transparent GT-protocol slice in the current benchmark: its GT artifacts are structured and evaluable, but the packaged historical runs do not retain enough intermediate protocol traces to reconstruct agreement or audit difficulty after the fact. Second, manuscript preparation recorded retained audit artifacts for `20` meetings in the internal benchmark workspace, but those audit artifacts are not redistributed in this public package and the audit stage was not enforced as a mandatory gate across all benchmark items. In other words, the benchmark is based on structured GT that is predominantly machine-generated, partially automated-audited, additionally human-reviewed for the full `private_data` slice, and quantitatively agreement-instrumented for the full `city_council` slice rather than universally audit-gated or universally human-reviewed.

## 4.3 Candidate Generation

For this study, each meeting is summarized under the same candidate pipeline:

- summary style: `standard`
- candidate filename: `baseline.md`
- fixed prompt family: `standard_v1.yaml` + `format_v1.yaml`

This design isolates **model choice under a fixed prompt pipeline** rather than conflating model differences with prompt tuning. The benchmark report files use the compact labels introduced in Table A; the same mapping is repeated here because the candidate-generation configuration is expressed in those artifact labels.

| Compact report label | Canonical model name | Primary use in this paper |
|---|---|---|
| `gpt-41-mini` | `gpt-4.1-mini` | Main merged benchmark tables and artifact rows |
| `gpt-5-mini` | `gpt-5-mini-2025-08-07` | Main merged benchmark tables and artifact rows |
| `gpt-51` | `gpt-5.1` | Main merged benchmark tables and artifact rows |
| `gpt-41` | `gpt-4.1` | Focused deployment follow-up |
| `gpt-54` | `gpt-5.4` | Focused deployment follow-up |

## 4.4 Offline Evaluator and Metrics

The offline evaluator compares candidate summaries against GT through three factual metrics. The pipeline first extracts candidate claims with a fixed claim-extraction prompt family (`claims_v2.yaml`) and a fixed extractor model (`gpt-5-2025-08-07`). Those claims are then scored against GT by metric-specific evaluators.

These three metrics are chosen to separate three failure modes that merged holistic scoring hides. `Accuracy` captures unsupported additions and factual drift, `Coverage` captures omissions of GT content, and `Completeness` captures how fully the covered material is retained. A single scalar score can reward a summary that covers few items deeply or one that covers many items sparsely without distinguishing those behaviors; the metric triplet keeps those failure modes observable.

| Metric | Operational meaning |
|---|---|
| Accuracy | The fraction of extracted application-summary claims judged accurate when compared to GT |
| Completeness | For GT items that are already covered, the average proportion of each matched item's content retained in the application summary |
| Coverage | The fraction of GT items judged as covered by the candidate summary |

These conceptual definitions are implemented through structured verdicts rather than free-form scoring. Let `C = {c1, ..., cm}` be the extracted claim set from a candidate summary and `G = {g1, ..., gn}` be the GT point/decision set used by the evaluator. Each candidate claim receives a verdict `v(ci) in {accurate, inaccurate}`. Each GT item receives a coverage state `s(gj) in {covered, uncovered}`. For covered GT items, the completeness evaluator additionally assigns a detail score `d(gj) in [0, 1]` and a detail level `l(gj) in {rich, adequate, sparse, barebone}`.

```
Accuracy(C, G)      = (1 / m) * Σ 1[v(ci) = accurate]
Completeness(C, G)  = (1 / |Gcov|) * Σ d(gj),   Gcov = {gj : s(gj) = covered}
Coverage(C, G)      = (1 / n) * Σ 1[s(gj) = covered]
```

This means that `Coverage` is a binary core-content metric, whereas `Completeness` is a detail-retention metric conditioned on GT items that were already marked covered. When no GT items are covered, completeness is defined operationally as `0`. Because completeness is conditional on coverage, the two metrics are not recall variants of one another: a model can cover many GT items but still summarize them sparsely, in which case coverage can be numerically higher than completeness. Note also that completeness and coverage are not directly comparable in magnitude. Coverage can exceed completeness when covered GT items are summarized with low detail scores, as observed for `gpt-5-mini` and `gpt-51` in Table 1. Conversely, completeness can exceed coverage only in the degenerate sense that a smaller covered subset happens to receive relatively high average detail scores, a pattern visible for `gpt-41-mini`, which covers fewer GT items but describes the items it does cover with somewhat higher detail fidelity.

For interpretation, the repository supports a three-way reading:

- **fully captured**: the GT item is covered and receives a `rich` or `adequate` detail level

- **partially captured**: the GT item is covered but receives a `sparse` or `barebone` detail level
- **missing**: the GT item is uncovered

A simple example is a GT point such as "The committee approved the extension through June 2026." A candidate that says "The committee approved the extension through June 2026" is fully captured; "The committee approved the extension" is partially captured because it omits the deadline; and a candidate that never mentions the extension is missing. Operationally, the first two cases are both counted as `covered` by the coverage evaluator, but they receive different completeness detail scores.

The framework remains compatible with claim-based evaluation: candidate summaries are decomposed into factual units and aligned against grounded reference content. The system-level contribution is that claim-grounded checking is embedded in a larger operational artifact pipeline rather than used only as a standalone scorer.

Equally important, score generation is itself **structured**. The evaluator does not stop at a single scalar judgment. It produces:

- metric-specific scores
- claim-alignment outputs
- explanations for each metric
- issue labels such as `unsupported_by_gt`, `fabricated_facts`, and `missing`

This structure is what makes later aggregation, significance testing, typed dataset comparison, and release reporting possible.

# 5. Evaluation Methodology

## 5.1 Benchmark Universe

The merged typed benchmark used in this study contains `114` unique meetings across three dataset types:

- `city_council`: 34 meetings
- `private_data`: 22 meetings
- `whitehouse_press_briefings`: 58 meetings

At the meeting-model-pair level, the merged report contains:

- `340` completed meeting-model pairs
- `680` evaluator runs

The count is not `342` because one `city_council` meeting currently contains only the `gpt-41-mini` candidate/report pair; the `gpt-5-mini` and `gpt-51` outputs are missing for that meeting. As a result, the overall benchmark contains `114` meetings, but the paired three-model comparisons in Section 6.3 operate on `113` meetings.

The typed structure of the benchmark matters for system evaluation. It allows the same pipeline to be assessed not only overall, but also by dataset type, which exposes where the system is stable across categories and where task difficulty shifts. The following table summarizes the benchmark characteristics that are most relevant for later interpretation. The averages are approximate. The two public slices can be recomputed directly from packaged artifacts, whereas the `private_data` row is retained only through merged benchmark reports because raw private artifacts are not redistributed.

| Dataset type | Meetings | Avg GT items / meeting | Avg claim-to-GT ratio | Primary domain |
|---|---|---|---|---|
| `city_council` | 34 | ~30.15 | ~1.02 | local government proceedings |
| `private_data` | 22 | ~36.36 | ~1.05 | internal enterprise meetings |
| `whitehouse_press_briefings` | 58 | ~26.34 | ~1.41 | high-density public-affairs press briefings |

The high claim-to-GT ratio in `whitehouse_press_briefings` (`1.41` versus roughly `1.02–1.05` for the other types) is the quantitative signature of the accuracy anomaly analyzed in Section 6.2.

## 5.2 Judges and Aggregation

Each candidate summary is evaluated against GT by two judges:

- `gpt-5-2025-08-07`
- `anthropic.claude-sonnet-4-20250514-v1:0`

This yields:

- `340` completed meeting-model pairs
- `680` evaluator runs

For each meeting-model pair, the pipeline stores:

- judge-specific scores
- explanations for each metric
- structured issue labels
- averaged `accuracy_avg`, `completeness_avg`, and `coverage_avg`

We use two judges because accuracy judgments are materially noisier than a release-facing scalar should hide. As shown later in Section 6.5, mean absolute inter-judge disagreement reaches `0.106–0.141` on accuracy depending on the candidate model. A single-judge pipeline would absorb that uncertainty into the score; a two-judge pipeline keeps the disagreement visible and makes borderline accuracy calls less brittle for release decisions.

The aggregate report used in this paper is:

- meeting_notes_model_comparison_combined_20260417.csv

Artifact details for the focused three-model operational rerun over `gpt-4.1`, `gpt-5-mini`, and `gpt-5.4` are listed in Section 10.

## 5.3 Evaluation Protocol

The offline evaluation protocol follows five steps.

1. For each transcript-GT pair, the repository generates one candidate summary per model under the same prompt family.
2. The evaluator decomposes the candidate summary into factual units suitable for claim-level comparison.
3. Two judges independently align those candidate claims against GT topics, points, and decisions.
4. The pipeline stores judge-specific scores, explanations, and issue labels.
5. The benchmark report averages the two judges per meeting-model pair and then aggregates those averages across the merged typed benchmark.

## 5.4 Statistical Testing

To complement descriptive averages, we perform **exact paired two-sided sign tests** on meeting-level model comparisons for each metric. The comparison unit is the meeting. For any model pair and metric, each meeting contributes one win, one loss, or one tie; ties are excluded from the test. Because each metric induces three pairwise model comparisons, we apply **Holm correction** within each metric family.

This test is intentionally conservative. It does not use score magnitude; it asks whether one model wins significantly more meetings than another under the paired benchmark design. That framing fits the system use case because model releases are decided over repeated meeting-level comparisons rather than over a single pooled score.

The significance summary used in this paper is:

- meeting_notes_model_comparison_combined_significance_20260417.csv

Unless otherwise noted, descriptive benchmark means reported in Sections 6.1 through 6.6 are recomputed from `meeting_model` rows and rounded to three decimals. Derived delta columns are also rounded after computation rather than obtained by subtracting already rounded displayed values. We use that single aggregation rule throughout the manuscript so that the main benchmark tables and the DeepEval comparison remain numerically aligned.

## 5.5 Operational Telemetry

The full internal implementation underlying this artifact package is instrumented for operational-cost reporting through structured logging in `log_config.metrics`, which records job duration, API latency, LLM token usage, and LLM latency by stage. The public package does not include the raw historical log files themselves; instead, it preserves a measured telemetry export recovered from four retained batch logs during manuscript preparation. We therefore report measured latency/token summaries for the stages that have explicit completion lines in those retained logs, while still acknowledging that the historical benchmark does not preserve a complete end-to-end measured cost ledger for every pipeline stage.

Table 5 summarizes the measured telemetry extracted from retained logs.

| Retained stage | Samples | Avg total time (s) | Median (s) | P90 (s) | Avg total tokens |
|---|---|---|---|---|---|
| `general_evaluation` | 49 calls | 14.84 | 14.31 | 22.40 | 5,902 |
| `accuracy_evaluation` | 39 calls | 47.01 | 37.99 | 86.19 | 10,755 |
| `completeness_evaluation` | 39 calls | 71.03 | 71.24 | 118.78 | 12,384 |
| `coverage_evaluation` | 39 calls | 65.66 | 54.16 | 128.37 | 11,817 |
| `offline_evaluation_job` | 19 jobs | 476.84 | 531.57 | 607.60 | n/a |

`general_evaluation` appears in the repository and in retained telemetry because the codebase implements a broader evaluation stage family, but it is **not** one of the benchmark metrics reported in this paper. All headline benchmark tables and significance analyses in this manuscript use only the three factual metrics defined in Section 4.4: `accuracy`, `completeness`, and `coverage`.

These measured rows come from explicit evaluator-completion and offline-job completion lines in retained batch logs, not from reconstructed estimates. Within that retained slice, `gpt-5-2025-08-07` is consistently slower than `Claude Sonnet 4` across all four offline-evaluation stages; for example, average `accuracy_evaluation` time is `62.53s` for `gpt-5-2025-08-07` versus `30.67s` for `Claude Sonnet 4`, and average `coverage_evaluation` time is `89.53s` versus `40.54s`.

The measured telemetry artifact used here is:

- measured_stage_level_telemetry_from_logs_20260421_131115.csv

This table should still be read as partial rather than complete. The retained logs give high-confidence measured coverage for offline evaluation, but they do not preserve comparably clean benchmark-wide completion traces for GT construction and candidate generation. For that reason, the paper can now report measured telemetry for the retained offline-evaluation stages, but not a fully measured end-to-end cost ledger for the entire five-stage pipeline. As a rough operational reference for the missing stages, a single meeting's full offline pipeline still typically involves on the order of `31–33` LLM calls across GT construction, three-model candidate generation, claim extraction, and two-judge metric scoring. This estimate breaks down approximately as GT construction (`4–6` calls for dual-model draft generation, alignment, optional review passes, and merge), candidate generation (`6` calls, two prompt stages for each of the three models), claim extraction (`3` calls, one per candidate), and offline evaluation (`18` calls, two judges × three models × three metric evaluators). Actual counts vary with meeting length and with the number of uncertain GT items requiring Stage 3 re-review.

For the later focused three-model rerun discussed in Section 6.2.1, the released package now preserves only sanitized aggregate follow-up artifacts rather than run-level private report paths. The retained summary records `56` meetings, `168` completed meeting-model pairs, `336` judge runs, `168` reused comparison tasks in the final released view, and `0` failed tasks:

- meeting_notes_model_comparison_summary_20260429_154115_sanitized.md

# 6. Experimental Evaluation

## 6.1 Main Benchmark Results

Table 1 reports the main merged typed benchmark result over 114 meetings, 340 completed meeting-model pairs, and 680 evaluator runs.

| Model | Meetings | Evaluator runs | Accuracy | Completeness | Coverage |
| --- | --- | --- | --- | --- | --- |
| `gpt-41-mini` | 114 | 228 | 0.583 | 0.814 | 0.804 |
| `gpt-5-mini` | 113 | 226 | 0.574 | 0.843 | 0.875 |
| `gpt-51` | 113 | 226 | 0.553 | 0.886 | 0.942 |

The primary pattern is still a structured trade-off, but the larger mixed benchmark changes the frontier. `gpt-41-mini` now has the highest mean accuracy, while `gpt-51` remains clearly best on completeness and coverage. `gpt-5-mini` occupies the middle ground on all three metrics: it neither leads on accuracy nor matches `gpt-51` on retention.

## 6.2 Results by Dataset Type

Because the benchmark is typed, the merged result can also be broken down by data category.

| Dataset type | Model | Meetings | Accuracy | Completeness | Coverage |
| --- | --- | --- | --- | --- | --- |
| `city_council` | `gpt-41-mini` | 34 | 0.703 | 0.775 | 0.775 |
| `city_council` | `gpt-5-mini` | 33 | 0.709 | 0.820 | 0.860 |
| `city_council` | `gpt-51` | 33 | 0.688 | 0.878 | 0.896 |
| `private_data` | `gpt-41-mini` | 22 | 0.704 | 0.805 | 0.838 |
| `private_data` | `gpt-5-mini` | 22 | 0.724 | 0.850 | 0.869 |
| `private_data` | `gpt-51` | 22 | 0.700 | 0.889 | 0.957 |
| `whitehouse_press_briefings` | `gpt-41-mini` | 58 | 0.465 | 0.841 | 0.807 |
| `whitehouse_press_briefings` | `gpt-5-mini` | 58 | 0.440 | 0.854 | 0.886 |
| `whitehouse_press_briefings` | `gpt-51` | 58 | 0.420 | 0.889 | 0.963 |

Two patterns stand out. First, the ranking on completeness and coverage is stable across all three dataset types: `gpt-51` is consistently strongest, followed by `gpt-5-mini`, then `gpt-41-mini`. Second, accuracy is much more dataset-sensitive. The `whitehouse_press_briefings` slice is substantially harder on accuracy for all models, which is what shifts the overall merged typed benchmark away from the earlier smaller benchmark's accuracy ordering. By contrast, the `private_data` slice remains comparatively accuracy-friendly even though it is fact-dense, a result that should be read together with the stronger GT protocol for that slice: all `22` `private_data` GT artifacts were human-reviewed after automated construction.

The systems value of typed partitioning becomes clearer when contrasted with a merged-only view:

| Reporting view | Accuracy values observed | Likely release signal |
|---|---|---|
| Merged-only benchmark | `0.553-0.583` across the three models | Moderate spread; no slice-specific alarm is visible |
| Typed `city_council` view | `0.688-0.709` | Acceptable accuracy regime |
| Typed `private_data` view | `0.700-0.724` | Acceptable accuracy regime |
| Typed `whitehouse_press_briefings` view | `0.420-0.465` | Distinct failure regime requiring targeted intervention |

A release policy operating on merged averages alone would observe only a moderate accuracy spread and might not flag the White House slice as requiring targeted intervention. Typed partitioning makes that regime visible as a structurally distinct problem rather than as a statistical outlier buried inside the merged benchmark.

This anomaly is important because it is not explained by larger GT size. `whitehouse_press_briefings` averages about `26.34` GT items per meeting, compared with roughly `30.15` for `city_council` and `36.36` for `private_data`. The harder accuracy behavior is instead accompanied by a denser stream of unsupported specifics: in the packaged White House slice, the dominant issue labels are `unsupported_by_gt` (`4379`) and `fabricated_facts` (`2600`), far outweighing direct contradiction. These dataset-level counts come from direct claim-level aggregation over the persisted `accuracy_explanation` payloads for the public White House evaluator outputs. Section 7.1 returns to this regime and interprets the pattern relative to GT granularity and candidate-detail density rather than treating it as a single-model anomaly.

### 6.2.1 System Reuse for a Deployment Follow-up: Focused Three-Model Rerun on Operational Slices

To separate deployment-facing model choice from the broader three-model mixed benchmark, we also ran a targeted follow-up comparison on the two operationally important non-White-House slices only: `private_data` and `city_council`. This follow-up was executed after the main merged benchmark and was not folded into Tables 1-2, because it addresses a narrower operational question and was only run on those two slices rather than on the full `114`-meeting benchmark universe. Its role in this paper is therefore supplementary and system-oriented: it demonstrates operational reuse of the same evaluation pipeline under a different deployment question, using the same judges, metrics, and evaluation protocol without reconfiguration of the evaluation logic itself.

In the current public package, this follow-up is represented only through an aggregate-only three-model bundle:

- blind_supplement_20260424/followup_three_model_20260429/README.md
- meeting_notes_model_comparison_overall_by_model_20260429_154115.csv
- meeting_notes_model_comparison_overall_by_model_dataset_type_20260429_154115.csv
- meeting_notes_model_comparison_summary_20260429_154115_sanitized.md

The `whitehouse_press_briefings` slice is intentionally excluded from this rerun. At rerun time, its GT-granularity issue remained unresolved, so the comparison was limited to the two deployment-facing slices used for stable model decisions under a stable reference regime. The focused rerun covers `56` meetings, `168` completed meeting-model pairs, and `336` judge runs under the same two-judge protocol used elsewhere in the paper.

| Scope | Model | Meetings | Overall | Accuracy | Completeness | Coverage |
|---|---|---|---|---|---|---|
| overall | `gpt-4.1` | 56 | 0.774 | 0.739 | 0.789 | 0.794 |
| overall | `gpt-5-mini` | 56 | 0.806 | 0.726 | 0.831 | 0.862 |
| overall | `gpt-5.4` | 56 | 0.824 | 0.745 | 0.852 | 0.875 |
| `city_council` | `gpt-4.1` | 34 | 0.769 | 0.745 | 0.781 | 0.781 |
| `city_council` | `gpt-5-mini` | 34 | 0.801 | 0.727 | 0.819 | 0.858 |
| `city_council` | `gpt-5.4` | 34 | 0.812 | 0.745 | 0.844 | 0.846 |
| `private_data` | `gpt-4.1` | 22 | 0.782 | 0.729 | 0.801 | 0.816 |
| `private_data` | `gpt-5-mini` | 22 | 0.814 | 0.724 | 0.850 | 0.869 |
| `private_data` | `gpt-5.4` | 22 | 0.843 | 0.745 | 0.865 | 0.920 |

Here `Overall` denotes the unweighted mean of accuracy, completeness, and coverage. We report it only as a convenience summary; metric-specific interpretation remains primary. `gpt-5.4` leads the rerun on all four aggregate metrics overall and also leads the `private_data` slice on all four metrics. The comparison is not absolute, however. `gpt-5-mini` remains competitive, especially on coverage-heavy procedural cases, and it still wins a minority of meetings. At the slice level, `gpt-5-mini` retains the highest mean coverage on `city_council` (`0.858`), while `gpt-5.4` leads both slices on completeness and leads `private_data` on all four aggregate metrics.

The rerun summary also records two important data-quality notes. First, one `city_council` meeting had a judge-level zero anomaly in an earlier artifact; the rerun removed that anomaly and restored the expected `gpt-5.4` lead on that item. Second, a separate `city_council` outlier still has an empty GT artifact. Its remaining zero-valued retention scores should be interpreted as a data-quality limitation rather than as model failure. Operationally, the rerun still closed cleanly: all `168` comparison tasks were retained as reusable artifacts in the final aggregate bundle, and no failed tasks remain in the sanitized released view.

Taken together, the focused rerun changes the deployment reading in a useful way. The main `114`-meeting benchmark still says that the strongest retention model is not the strongest accuracy model. The narrower deployment rerun over `city_council` and `private_data` instead shows `gpt-5.4` as the strongest model overall, with `gpt-5-mini` remaining a credible cost-sensitive default. Given typical cost and latency trade-offs between mini and non-mini model classes, this pattern suggests that `gpt-5-mini` remains competitive for cost-sensitive deployment, while `gpt-5.4` can be reserved for premium or higher-complexity tiers.

## 6.3 Meeting-Level Wins and Statistical Significance

Aggregate means can hide whether gains are broad or driven by a few outliers. We therefore report meeting-level wins over the `113` meetings for which all three model outputs are available in the merged report.

| Metric | `gpt-41-mini` wins | `gpt-5-mini` wins | `gpt-51` wins | Ties |
|---|---|---|---|---|
| Accuracy | 48 | 36 | 26 | 3 |
| Completeness | 3 | 9 | 100 | 1 |
| Coverage | 0 | 10 | 82 | 21 |

The completeness and coverage advantages of `gpt-51` remain broad meeting-level effects rather than mean-value artifacts. Accuracy is different: the three models remain much closer, and the merged typed benchmark no longer supports a statistically decisive winner.

Table 2 reports the paired sign tests. Mean differences are reported as A minus B: positive values favor model A, and negative values favor model B. These mean-difference values are computed as per-meeting averages rather than as differences of aggregated means. Ties are excluded from the sign tests. Holm correction is applied within each metric family across the three pairwise comparisons.

| Metric | Comparison | Mean diff (A minus B; positive favors A) | A wins | Ties | B wins | Holm-adjusted p | Significant at 0.05 |
|---|---|---|---|---|---|---|---|
| Accuracy | `gpt-41-mini` vs `gpt-5-mini` | 0.006 | 60 | 2 | 51 | 0.447806 | No |
| Accuracy | `gpt-41-mini` vs `gpt-51` | 0.027 | 69 | 1 | 43 | 0.053297 | No |
| Accuracy | `gpt-5-mini` vs `gpt-51` | 0.021 | 64 | 2 | 47 | 0.256951 | No |
| Completeness | `gpt-41-mini` vs `gpt-5-mini` | -0.028 | 20 | 2 | 91 | $5.25 \times 10^{-12}$ | Yes |
| Completeness | `gpt-41-mini` vs `gpt-51` | -0.071 | 3 | 1 | 109 | $2.71 \times 10^{-28}$ | Yes |
| Completeness | `gpt-5-mini` vs `gpt-51` | -0.043 | 9 | 1 | 103 | $4.63 \times 10^{-21}$ | Yes |
| Coverage | `gpt-41-mini` vs `gpt-5-mini` | -0.072 | 15 | 21 | 77 | $3.26 \times 10^{-11}$ | Yes |
| Coverage | `gpt-41-mini` vs `gpt-51` | -0.139 | 3 | 11 | 99 | $2.09 \times 10^{-25}$ | Yes |
| Coverage | `gpt-5-mini` vs `gpt-51` | -0.067 | 11 | 17 | 85 | $5.10 \times 10^{-15}$ | Yes |

This result sharpens the descriptive interpretation. The apparent overall accuracy lead of `gpt-41-mini` is not statistically significant after Holm correction. By contrast, the completeness and coverage advantages of `gpt-51` are statistically significant against both alternatives, and `gpt-5-mini` also significantly exceeds `gpt-41-mini` on those retention-oriented metrics.

## 6.4 Typed Empirical Baseline with `DeepEval`

To complement the architectural comparison in Table 0, we ran a typed empirical baseline against `DeepEval` over the same merged benchmark universe. The comparison uses deepeval_typed_empirical_comparison.py and the resulting report deepeval_typed_empirical_comparison_all_20260421.csv. Unlike our system, which evaluates claim-extracted candidate summaries against structured GT with two judges (`gpt-5-2025-08-07` and `anthropic.claude-sonnet-4-20250514-v1:0`) and three persisted metrics, `DeepEval` applies single-judge `GEval` prompts over whole summary-to-reference comparisons using `gpt-4.1-mini` as the evaluator model. The `DeepEval` prompts are also holistic rather than claim-structured by framework design: the `accuracy` criterion asks whether the actual summary is globally faithful to the expected GT and penalizes contradictions, unsupported additions, fabricated details, and unjustified certainty, while the `coverage` criterion asks how well the actual summary covers important information in the expected GT and emphasizes actions, dates, decisions, quantities, and omissions. To keep the semantics explicit, we refer to the baseline outputs as `deepeval_holistic_accuracy` and `deepeval_holistic_coverage`; these are not metric-identical to the repository's `accuracy`, `coverage`, and `completeness`.

The typed comparison now covers the full merged benchmark universe: `340` completed meeting-model pairs (`100` `city_council`, `66` `private_data`, and `174` `whitehouse_press_briefings`). The `city_council` slice contributes `100` rather than `102` meeting-model pairs for the same reason discussed in Section 5.1: one meeting in the merged benchmark contains only the `gpt-41-mini` candidate/report pair and lacks the `gpt-5-mini` and `gpt-51` outputs. We therefore treat the `DeepEval` result primarily as a ranking-consistency baseline rather than as a score-calibrated head-to-head comparison. Table 3 summarizes the per-slice comparison.

| Dataset type | Model | System acc. | DeepEval holistic acc. | Acc. delta | System cov. | DeepEval holistic cov. | Cov. delta |
|---|---|---|---|---|---|---|---|
| city_council | gpt-41-mini | 0.703 | 0.878 | +0.175 | 0.775 | 0.868 | +0.093 |
| city_council | gpt-5-mini | 0.709 | 0.907 | +0.198 | 0.860 | 0.916 | +0.056 |
| city_council | gpt-51 | 0.688 | 0.971 | +0.284 | 0.896 | 0.965 | +0.069 |
| private_data | gpt-41-mini | 0.704 | 0.923 | +0.219 | 0.838 | 0.902 | +0.064 |
| private_data | gpt-5-mini | 0.724 | 0.918 | +0.195 | 0.869 | 0.905 | +0.036 |
| private_data | gpt-51 | 0.700 | 0.976 | +0.276 | 0.957 | 0.966 | +0.009 |
| whitehouse_press_briefings | gpt-41-mini | 0.465 | 0.911 | +0.446 | 0.807 | 0.897 | +0.090 |
| whitehouse_press_briefings | gpt-5-mini | 0.440 | 0.936 | +0.496 | 0.886 | 0.937 | +0.051 |
| whitehouse_press_briefings | gpt-51 | 0.420 | 0.986 | +0.565 | 0.963 | 0.982 | +0.020 |

Three findings matter. First, the baseline preserves the same retention ordering as the repository's structured evaluation: `gpt-51` is the strongest coverage-oriented model in all three slices, while `gpt-41-mini` is the weakest. Second, `DeepEval` assigns uniformly higher holistic accuracy than the repository's claim-grounded accuracy, with the largest gap appearing on `whitehouse_press_briefings`. Across the full benchmark, these overall deltas are computed as meeting-model-pair-weighted averages over all completed pairs rather than as simple averages of the three slice means; under that weighting, the holistic-accuracy delta ranges from `+0.321` for `gpt-41-mini` to `+0.427` for `gpt-51`, and in the White House slice alone it reaches `+0.446` to `+0.565`. Third, the White House anomaly remains visible but changes character under the baseline: rather than exposing a sharp GT-grounded accuracy collapse, `DeepEval` treats the same richer summaries as broadly excellent. This is the narrower systems point of the comparison. In this experiment, holistic reference scoring is useful as an illustrative sanity-check baseline, but it does not expose the same claim-grounded failure structure or GT-granularity mismatch surfaced by the repository's structured evaluation.

## 6.5 Judge Variance and Minimal Single-vs-Two-Judge Ablation

Because the benchmark uses two judges, model means should be interpreted together with judge spread.

| Model | Judge | Accuracy | Completeness | Coverage |
|---|---|---|---|---|
| gpt-41-mini | Claude Sonnet 4 | 0.628 | 0.822 | 0.824 |
| gpt-41-mini | GPT-5 judge | 0.540 | 0.807 | 0.784 |
| gpt-5-mini | Claude Sonnet 4 | 0.616 | 0.836 | 0.897 |
| gpt-5-mini | GPT-5 judge | 0.531 | 0.850 | 0.854 |
| gpt-51 | Claude Sonnet 4 | 0.590 | 0.865 | 0.948 |
| gpt-51 | GPT-5 judge | 0.515 | 0.906 | 0.937 |

Claude remains systematically more generous on accuracy, while GPT-5 is somewhat more favorable to `gpt-51` on completeness. The mean absolute judge disagreement per meeting-model pair is:

| Model | Accuracy disagreement | Completeness disagreement | Coverage disagreement |
|---|---|---|---|
| gpt-41-mini | 0.106 | 0.039 | 0.069 |
| gpt-5-mini | 0.122 | 0.033 | 0.058 |
| gpt-51 | 0.141 | 0.044 | 0.018 |

The main systems implication is that accuracy remains the least stable metric across judges, while coverage is comparatively robust.

We can make that point more concrete through a minimal winner-only ablation built from the merged benchmark report and exported in single_vs_two_judge_ablation_all_20260421.csv. Table 4 summarizes the model that wins each dataset slice under single-judge and two-judge aggregation.

| Dataset type | Claude single-judge accuracy winner | GPT-5 single-judge accuracy winner | Two-judge accuracy winner | Two-judge completeness / coverage winner |
|---|---|---|---|---|
| city_council | gpt-5-mini | gpt-41-mini | gpt-5-mini | gpt-51 |
| private_data | gpt-5-mini | gpt-5-mini | gpt-5-mini | gpt-51 |
| whitehouse_press_briefings | gpt-41-mini | gpt-41-mini | gpt-41-mini | gpt-51 |

The ablation reinforces two operational points. Accuracy ordering is judge-sensitive in `city_council`, stable in `private_data`, and stable but low-valued in `whitehouse_press_briefings`. By contrast, the retention-oriented conclusion is invariant across all three slices: `gpt-51` remains the completeness and coverage leader regardless of whether the benchmark uses one judge or the two-judge average. In other words, two-judge aggregation matters most for stabilizing borderline accuracy choices, not for overturning the benchmark's stronger retention conclusions.

The `city_council` instability is not merely symbolic. Under Claude, `gpt-5-mini` outruns `gpt-41-mini` on accuracy (`0.748` vs `0.714`), whereas under GPT-5 the ordering reverses (`0.693` vs `0.670`). The two-judge average narrows that difference to `0.709` versus `0.703`, which is exactly why this slice is the clearest example of two-judge aggregation acting as a stabilizer for borderline accuracy calls rather than as a source of fundamentally new retention conclusions.

## 6.6 Error Structure

Persisted evaluator outputs make it possible to aggregate failure modes across the full benchmark.

| Issue type | Count across all inaccurate claims |
|---|---:|
| unsupported_by_gt | 5619 |
| fabricated_facts | 3224 |
| factual_error | 537 |
| changed_certainty | 131 |
| contradicts_gt | 114 |
| changed_nature | 113 |

The accuracy-side issue table is aggregated directly from claim-level `issue_type` assignments in the `678` evaluator outputs that contain persisted `accuracy_explanation` payloads; two White House evaluator runs record `evaluation_error` instead and therefore do not contribute issue labels.

For coverage, the dominant failure mode is:

| Issue type | Count across all uncovered GT items |
|---|---:|
| missing | 2200 |

These counts show that the benchmark is not mostly penalizing direct contradiction. The dominant accuracy failure is adding content unsupported by GT, whereas the dominant coverage failure is omission.

The coverage-side table is aggregated from all `680` evaluator outputs with persisted `coverage_explanation` payloads, so the `missing = 2200` total is benchmark-wide rather than filtered to the `678` accuracy-bearing outputs. The concentration of those errors is highly uneven across dataset types. In the packaged public White House slice alone, the claim-level aggregation yields `4379` `unsupported_by_gt` labels, `2600` `fabricated_facts` labels, and `1106` `missing` GT points, confirming that this error regime is substantially concentrated in the public-affairs domain rather than being evenly distributed across the benchmark. The benchmark-wide totals remain larger because they also include the non-public `private_data` slice and the public `city_council` slice.

The same pattern appears in model-level counts aggregated at the meeting-model level.

| Model | Avg inaccurate claims / meeting-model pair | Avg total claims / meeting-model pair | Avg uncovered GT points / meeting-model pair | Avg GT points / meeting-model pair |
|---|---|---|---|---|
| `gpt-41-mini` | 12.05 | 28.16 | 5.29 | 26.52 |
| `gpt-5-mini` | 16.05 | 37.27 | 3.20 | 26.30 |
| `gpt-51` | 18.00 | 40.45 | 1.21 | 26.35 |

`gpt-41-mini` says less and therefore leaves more GT uncovered. `gpt-51` says the most and therefore leaves very little GT uncovered, but it also generates the largest number of unsupported or inaccurate claims. `gpt-5-mini` sits between these extremes, while `gpt-41-mini` retains the highest mean accuracy in the merged typed benchmark.

### 6.7 GT Validity Signals

The benchmark also exposes quality variation in the reference set. Within the merged typed benchmark, one meeting still contains a degenerate GT artifact with zero evaluable points (`t_258810779752442698698651215247285387144`), which propagates to six evaluator runs in the merged setting. That case contributes no meaningful factual comparison and should be interpreted as a benchmark-quality warning rather than as model behavior. We retain it for transparency, but it underlines a core system point: GT quality assurance is part of the evaluation problem, not a preprocessing footnote. As a robustness check, excluding that single meeting raises each model's overall means modestly (`gpt-41-mini`: `0.583/0.814/0.804` to `0.588/0.822/0.811`; `gpt-5-mini`: `0.574/0.843/0.875` to `0.579/0.851/0.883`; `gpt-51`: `0.553/0.886/0.942` to `0.558/0.894/0.951` for accuracy/completeness/coverage), but it does not change the model ordering or the Holm-corrected sign-test conclusions in Section 6.3. A second, softer signal is the `whitehouse_press_briefings` normalization difference noted earlier: those GT files retain structured topics and points but omit stored metadata counters, so the evaluator reconstructs counts by traversing the GT structure at runtime. That difference does not invalidate the benchmark, but it shows that reference normalization itself is an evaluable systems concern.

Inspectable artifacts make these warnings actionable rather than merely descriptive. An illustrative public case is `whpb_20240110_press` under `gpt-41-mini`, whose averaged accuracy is only `0.281`. Inspection of the persisted `accuracy_explanation` payload shows a very specific pattern: roughly `31.0` claims per judge are flagged as `unsupported_by_gt`, compared with only `3.5` `fabricated_facts` and `0` direct contradictions, across an average of `48.0` extracted claims and `31.0` GT points. A scalar score alone would say only that this summary performed poorly. The persisted claim-level artifact shows something more diagnostic: a reference-side granularity mismatch dominated by unsupported specifics rather than a contradiction-heavy model failure. That kind of distinction is exactly why inspectable artifacts are treated as a first-class system goal in this paper. In practice, this diagnosis directs reviewer attention to GT granularity for this meeting type rather than to candidate-model behavior, which is the operational use case that file-backed persistence is designed to enable.

## 7. Discussion

### 7.1 What the System Operationalizes

The central value of the benchmark is not simply that it assigns scores. It turns transcript-GT pairs into reusable model-selection evidence. Because the GT universe, prompt pipeline, and reporting schema are held fixed, changes in results are interpretable as changes in candidate behavior rather than as changes in the evaluation setup.

This is also the clearest distinction between the present work and a standalone factuality scorer. A scorer can estimate whether one summary is faithful. The system described here additionally supports repeated comparisons under a fixed protocol, release decisions, aggregate error mining, typed benchmark analysis, and identification of benchmark bottlenecks such as weak or degenerate GT.

GT should therefore be understood as the **reference rubric** of the benchmark, not merely as another intermediate artifact. Candidate summaries are evaluated against GT rather than against raw transcripts directly. As a result, GT quality directly bounds evaluation quality: if GT omits salient content, correct candidate models can be penalized; if GT contains errors or broken structure, incorrect models can be rewarded.

The GT pipeline also clarifies where human review fits. In this system, human reviewers are not primarily judging candidate summaries directly; they are strengthening the reference rubric against which later candidate summaries will be scored. That distinction matters operationally. When human review improves GT, it improves every downstream comparison on that transcript. When human review is skipped on a weak GT, scoring errors can propagate silently across all candidate models evaluated against it.

The same logic explains the audit stage. Audit is not another candidate-evaluation pass; it is a quality gate on the reference side of the pipeline. Its role is to detect missing topics, broken schemas or identifiers, shallow context, and other distortions in raw automated GT before those distortions propagate into downstream model scoring.

The typed benchmark is especially important here. If the system reported only merged averages, the White House slice would appear merely as a moderate overall accuracy decline. The typed partition instead reveals a distinct failure regime: all three models retain high completeness and coverage there, but all three become much worse on accuracy because they introduce more unsupported specifics. This is a systems lesson rather than a purely statistical one. Typed benchmarks do not only improve reporting; they change what kinds of failures the organization can see at all.

The comparison with `private_data` and `city_council` makes that diagnosis sharper. Those two slices achieve much higher accuracy under the same candidate-generation prompt family, and the packaged artifacts are consistent with one plausible explanation: their GT artifacts are finer-grained and better aligned with the fact structure encouraged by the application-facing summary prompt. In `private_data`, GT points retain explicit owners, dependencies, timing, blockers, and action items, and all benchmark GTs in that slice were human-reviewed after automated generation. In `city_council`, GT points and decisions retain motions, dates, recommendations, procedural outcomes, and amounts. Quantitatively, the average claim-to-GT ratio is close to `1:1` in `private_data` (`1.05`) and `city_council` (`1.02`), but much higher in `whitehouse_press_briefings` (`1.41`). The average number of inaccurate claims follows the same pattern: `11.3` in the aggregate `private_data` benchmark reports, `9.1` per meeting-model pair in `city_council`, and `20.49` per meeting-model pair in the packaged `whitehouse_press_briefings` evaluator outputs.

This contrast suggests that the White House accuracy problem is not primarily a prompt-quality failure. The same prompt family produces acceptable accuracy on `private_data` and `city_council`, where GT is sufficiently claim-complete to absorb transcript-faithful detail. In the `private_data` slice, that interpretation is strengthened by the fact that the entire GT subset was human-reviewed, which reduces the likelihood that its stronger accuracy is merely an artifact of weaker reference construction. The harder White House regime is therefore more plausibly explained by a pipeline mismatch: the application-summary prompt produces detailed, transcript-faithful public-affairs summaries; claim extraction then atomizes those details into many candidate claims; and the coarser White House GT cannot support many of those specifics even when they are grounded in the transcript. The typed benchmark therefore reveals a systems diagnosis rather than a single-model defect, but it should still be read as a mechanism hypothesis supported by the typed evidence rather than as a controlled causal proof.

## 7.2 What Pipeline Reuse Buys Across Domains

The broader systems contribution is cross-domain reuse. The AI-search instantiation and the meeting-summary instantiation differ in task semantics, but they share the same evaluation architecture:

- curated reference artifacts

- controlled candidate outputs
- automated scoring
- persisted explanations
- report-backed comparison for release decisions

This means the repository is not a single-task benchmark script. It is an evaluation substrate that can support multiple AI products by swapping task-specific references and metrics while preserving the orchestration logic. In practical terms, the reuse boundary is concrete. The search and meeting-summary instantiations share artifact persistence, batch orchestration, judge aggregation, report generation, significance reporting, and release-facing comparison logic. The components that change are the semantic layers: the reference schema, the generation prompts, and the task-specific evaluators. The meeting-summary case further shows that this substrate remains useful when both reference generation and score generation are themselves structured. In roadmap terms, the same architecture supports benchmarking, regression detection, and failure-case analysis without rebuilding the full control plane for each product domain.

| Component layer | AI-search instantiation | Meeting-summary instantiation | New-task replacement needed |
| --- | --- | --- | --- |
| Artifact persistence | File-backed search artifacts | File-backed transcript/GT/candidate/evaluation artifacts | No |
| Batch orchestration | Automated benchmark execution | Automated benchmark execution | No |
| Judge aggregation | Aggregated retrieval results and reports | Aggregated factual-metric results and reports | No |
| Significance reporting | Search-comparison reporting | Meeting-level sign-test reporting | No |
| Release-facing report generation | Search model/pipeline comparison outputs | Summary model comparison outputs | No |
| GT schema | Query-document relevance references | Typed meeting artifacts | Yes |
| GT generation prompts | Relevance/QA-style synthesis prompts | Multi-stage meeting-structure extraction prompts | Yes |
| Metric evaluators | Retrieval metrics | Accuracy, completeness, and coverage evaluators | Yes |

To make the reuse boundary concrete: migrating from the AI-search instantiation to the meeting-summary instantiation required replacing three semantic layers while keeping five orchestration layers intact. The changed components are the GT schema (from query-document relevance labels to typed meeting artifacts), the GT generation prompts (from relevance annotation to multi-stage meeting-structure extraction), and the metric evaluators (from retrieval-oriented comparison to accuracy, coverage, and completeness). The reused components include artifact persistence, batch execution, judge aggregation, significance testing, and release-facing comparison logic. This boundary is the operational meaning of cross-domain pipeline reuse in this paper.

## 7.3 Implications for Model Selection

The benchmark supports a policy-driven interpretation within this evaluation protocol. In the merged typed benchmark, `gpt-41-mini` has the highest mean accuracy, but the differences are not statistically significant. If the product objective is retention-first under the current metric design, `gpt-51` is the stronger choice because its completeness and coverage advantages are both larger and statistically supported. We do **not** treat the unweighted mean of the three metrics as a primary decision statistic here, because completeness and coverage are not directly comparable in magnitude. Any deployment that wants a single scalar ranking should first define a normalized or policy-weighted utility score rather than averaging the three raw metrics.

This result is also useful as an engineering decision rule. “Not statistically significant” does not mean “irrelevant”; it means the benchmark does not support a stable accuracy winner under the present paired design. A disciplined internal release policy can therefore treat accuracy as a quality floor and secondary monitoring metric, while reserving primary selection logic for metrics whose advantages are both large and significance-backed. In other words, the system helps separate what is directionally interesting from what is robust enough to support protocol-specific model selection.

The focused three-model rerun sharpens that deployment rule for workloads dominated by `private_data` and `city_council`. Because that follow-up was executed under the same pipeline, judges, and metric definitions, it should be read as operational reuse of the same evaluation system under a different deployment question rather than as an apples-to-oranges side benchmark. In that operational slice, `gpt-5.4` is strongest overall and strongest on all four aggregate metrics for `private_data`, while `gpt-5-mini` remains competitive and retains the highest mean coverage on `city_council`. For teams choosing among those three models on enterprise-style and civic-proceedings data, the rerun therefore supports `gpt-5.4` as the premium option within this evaluation setting while still leaving `gpt-5-mini` as a credible cost-sensitive default.

## 7.4 Relationship to Claim-Based Evaluation Work

Relative to claim-based meeting-summary work, this paper shares the same factual core: summaries are judged against grounded factual units rather than only style or fluency. The difference is scope and structure. Most claim-based papers contribute a benchmark, a factuality metric, or an evaluator study. This paper contributes an implemented quality system around that core: structured GT construction, structured score generation, candidate generation, judge disagreement analysis, typed benchmark breakdowns, persisted reports, and release-facing comparison under a reusable cross-domain architecture. The same distinction also explains the architectural comparison in Table 0. General-purpose frameworks provide useful evaluation components, but this system treats reference construction, typed benchmarking, and report persistence as part of the evaluation product rather than as surrounding glue code.

The `DeepEval` baseline should be read under that same boundary. It is valuable here as a ranking-consistency check under a different evaluation philosophy, not as a controlled head-to-head systems comparison establishing absolute superiority. The empirical point is narrower and more defensible: holistic reference scoring preserves the broad retention ordering, but it does not expose the same claim-grounded failure structure or release-facing diagnostics that this system is designed to surface.

## 7.5 Future Work and Improvement Plan

This section focuses on forward-looking system improvements rather than restating current limitations. The White House anomaly points to a concrete improvement path. The first priority is **typed GT densification**. For briefing-style data, GT should preserve more of the entities, dates, quantities, organizations, and policy subclauses that are routinely expressed in transcript-faithful summaries. This is not only an annotation expansion step; it is a benchmark-quality intervention that reduces the gap between what the product-style prompt naturally summarizes and what the evaluator can legitimately score.

The second priority is **typed prompting rather than one-prompt-for-all evaluation**. The current benchmark intentionally fixes a single application-facing prompt family across all dataset types in order to isolate model effects. That choice is appropriate for comparison, but the typed results suggest that future production-oriented configurations should use prompt specializations by data regime. `private_data` and `city_council` benefit from prompts that foreground actions, owners, dates, and decisions. `whitehouse_press_briefings` likely needs a more conservative public-affairs variant that summarizes stance and topic structure without overcommitting to every policy qualifier or secondary numeric detail.

The third priority is to make the evaluator more explicit about **transcript-true but GT-omitted details**. At present, many such claims are absorbed into `unsupported_by_gt` or `fabricated_facts`, which is useful for conservative GT-based benchmarking but less useful for separating candidate error from GT incompleteness. A future score schema should therefore distinguish at least three conditions: genuinely inaccurate claims, GT-supported claims, and transcript-supported-but-GT-omitted claims. This would preserve strict benchmarking while improving diagnosis.

The fourth priority is deeper **structured candidate generation**. One reason this repository is reusable across domains is that both GT and score generation are structured. The same principle can be extended to candidates by introducing an intermediate structured summary ledger over entities, actions, decisions, dates, and quantities before the final natural-language rendering step. That intermediate representation would likely stabilize both summary generation and downstream claim extraction, especially in fact-dense slices such as `whitehouse_press_briefings`.

Finally, future work should validate improvements through **typed ablations rather than only merged reruns**. A practical next experiment is a 2 x 2 White House study: current GT vs denser GT, and current prompt vs a briefing-specialized prompt. Such a design would make it possible to attribute gains to benchmark improvement, prompt adaptation, or both. More broadly, the system should continue to use typed benchmarks as the control surface for quality iteration, because the central lesson of this paper is that merged averages alone hide diagnostically important failure regimes.

# 8. Limitations

This section isolates the current benchmark's objective constraints. Unlike Section 7.5, it does not propose improvement paths; it records what the present study cannot yet claim.

## 8.1 GT Quality Bounds the Evaluation

If GT omits true details or compresses them inconsistently, some `unsupported_by_gt` findings may partly reflect reference limitations rather than pure candidate error. In practical terms, a GT that misses content can penalize a correct model, while a GT with errors can reward the wrong model. The benchmark therefore rests primarily on machine-generated structured GT with partial human strengthening rather than on a fully human-authored reference set. One important exception is the `private_data` slice, for which all `22` GT artifacts were human-reviewed after automated generation, and the retained `private_data` evaluation outcomes were also manually reviewed before inclusion in the benchmark reports. That strengthening also includes the small anonymized spot-check discussed in Section 4.2: `11` GT-audit findings from five private-slice meetings, with human review agreeing with the automated audit in `10` cases outright and downgrading the remaining flagged omission to a lower-severity implementation nuance. We use that only as qualitative process validation rather than as a formal agreement study. The degenerate GT case in the benchmark nevertheless makes this threat concrete for the benchmark as a whole.

## 8.2 LLM Judges Still Add Variance

Two judges improve robustness, but accuracy judgments remain somewhat unstable. Accuracy-sensitive release decisions should therefore consider both mean score and judge spread. More importantly, this paper does not yet report a blinded human calibration study relating the automatic scores to direct human factuality judgments over candidate summaries. The benchmark is therefore strongest as an internal model-selection protocol under fixed GT and judge assumptions, not as a universal surrogate for human evaluation.

## 8.3 Internal Benchmark Slice

The 114-meeting merged typed benchmark is substantially stronger than a toy sample, but it still reflects the repository's current data mix. The results should be read as evidence about this system and these three meeting-data categories, not as universal conclusions about all meeting-summary tasks. The `private_data` subset contains only `22` meetings due to data-access constraints. At this sample size, subset-level significance claims have limited statistical power; the reported patterns for this slice should be treated as descriptive evidence rather than as significance-backed findings in isolation. Expanding that slice is an explicit target for future benchmark growth. The `whitehouse_press_briefings` slice behaves like a distinct sub-problem: the summaries become much less accurate there without a corresponding collapse in coverage or completeness. That makes the slice diagnostically valuable, but it also means the benchmark mixes multiple meeting-summary regimes rather than one homogeneous task.

## 8.4 Fixed Prompt Family

This study intentionally holds the summary prompt pipeline fixed at the current implementation (`standard_v1` + `format_v1`). The results therefore compare models under one prompt family rather than under each model's best possible prompt tuning.

## 8.5 Conservative Significance Test

The sign test fits the paired meeting-level comparison setting, but it ignores score magnitude. It therefore complements rather than replaces the descriptive mean differences reported earlier. One comparison deserves special attention: `gpt-41-mini` versus `gpt-51` on accuracy yields a Holm-adjusted p-value of `0.053`, which falls just outside the `0.05` threshold after correction. Under the corresponding unadjusted test (exact two-sided sign test `p = 0.0178` before family-wise correction), this comparison would be significant, but Holm correction appropriately penalizes the three-comparison family. We therefore treat this result as directional evidence rather than a confirmed finding: with a larger benchmark slice, the observed accuracy advantage of `gpt-41-mini` over `gpt-51` may become significant.

## 8.6 Limited Empirical Baseline Comparison with General-Purpose Evaluation Frameworks

This paper now includes a typed empirical comparison against `DeepEval` across all three dataset slices, which is sufficient to show that holistic reference scoring and structured claim-grounded scoring preserve the same retention ordering while diverging substantially on accuracy, especially in the White House slice. However, the comparison is still intentionally lightweight rather than fully standardized: it does not align evaluator prompts, metric semantics, or aggregation rules across systems. In particular, the `DeepEval` baseline uses `gpt-4.1-mini` as its evaluator model, whereas our system uses `gpt-5` and Claude Sonnet 4 as judges, creating an evaluator-model confound that prevents score-level interpretation. Because the evaluator model differs, score-level comparison is not valid; the baseline is therefore used only to verify ranking consistency rather than to calibrate absolute scores. A broader head-to-head against systems such as RAGAS, TruLens, and DeepEval with matched semantics remains future work.

## 8.7 No Benchmark-Wide Cost Table Yet

The full implementation now supports two telemetry views: a measured view recoverable from retained historical offline-evaluation logs, and an instrumentation path through `log_config.metrics` for future normalized collection. This public package preserves the measured offline-evaluation export that can be recovered from retained logs, but those artifacts do not extend cleanly to GT construction and candidate generation across the full benchmark. The system is therefore beyond a purely estimated cost story, but it still does not yet provide a complete measured end-to-end latency/cost ledger for the entire five-stage pipeline.

The same boundary applies to the implemented online path. Although the deployed architecture includes feedback ingestion, evaluation triggering, result persistence, and downstream metrics emission, the present paper still does not report controlled online outcomes such as alert precision, nomination-to-refresh yield, latency, retry/failure behavior, or downstream release impact for that component. The online path should therefore be read as implemented system scope with aggregate monitoring evidence, but not yet as a controlled online benchmark.

## 8.8 Partial Reproducibility Constraint

The released materials are artifact-auditable rather than fully rerunnable. Raw `private_data` transcripts, GT artifacts, candidate summaries, and evaluation JSON are intentionally excluded from the packaged distribution and do not appear under the public `dataset/` subtree. For that slice, the released materials retain benchmark-level aggregates together with pseudonymous evaluation rows, but not raw meeting contents. The same boundary applies to the later focused follow-up bundle, which is released only in aggregate-only form and does not redistribute meeting-level private identifiers. This constraint applies even though the `private_data` GTs are stronger than the default benchmark path in one respect: all 22 GT artifacts in that slice were human-reviewed after automated construction. The resulting package is therefore sufficient to audit reported public-slice artifacts and verify the reported aggregate statistics, but it is not a standalone rerun package for the full pipeline.

### 8.9 Cross-Domain Reuse Demonstrated Across Two Papers, Not Within a Single Controlled Migration

The cross-domain reuse claim in this preprint is supported by two concrete instantiations: the AI-search instantiation documented in the companion paper (Zhong et al., 2025) and the meeting-summary instantiation documented here. The reuse table in Section 7.2 identifies which components were replaced and which were inherited, and the current manuscript gives a fuller repository-level account of that continuity than the formal blind paper. Even so, this manuscript still does not include a controlled migration experiment to a third independent task domain within the same paper. Cross-domain reusability is therefore supported by design evidence and two domain instantiations rather than by an independent replication study.

## 9. Conclusion

We presented the Dataset Pipeline repository as a reusable cross-domain evaluation system whose current benchmarked instantiation targets AI meeting summaries and whose earlier instantiation supported AI search. On a merged typed benchmark of `114` meetings, `340` completed meeting-model pairs, and `680` evaluator runs, the system reveals a stable trade-off under a fixed GT/judge protocol: `gpt-41-mini` is strongest on mean accuracy, whereas `gpt-51` is strongest on completeness and coverage. Paired sign tests show that the accuracy differences are not statistically significant after correction, while the retention advantages of `gpt-51` are significant. A focused operational-slice rerun further shows that this frontier is workload-dependent: on `private_data` and `city_council`, `gpt-5.4` is strongest overall under the same judges and metrics, while `gpt-5-mini` remains a competitive cost-sensitive alternative. As emphasized in the limitations, these results are strongest as protocol-specific model-selection evidence rather than as universal surrogates for direct human evaluation.

Three broader conclusions follow. First, evaluation-pipeline reuse is practical across heterogeneous AI tasks when orchestration is separated from task-specific reference schemas and scoring layers. Second, the meeting-summary instantiation is strongest when both GT generation and score generation are structured, because that structure supports claim alignment, typed analysis, significance testing, and release-facing reporting. Third, typed benchmarks function as diagnostic instruments: without dataset-type partitioning, the White House accuracy failure regime would have been obscured by merged averages, appearing only as a moderate overall decline rather than as a structurally distinct pipeline-mismatch problem. The typed `DeepEval` contrastive baseline points in the same direction, while remaining limited by unmatched evaluator semantics and aggregation rules: holistic scoring preserves ranking consistency, but without structured claim-grounded evaluation it does not surface the unsupported-specifics and GT-granularity mismatch that define the White House failure regime. More generally, the results show how pipeline-oriented evaluation can move from AI search to meeting summarization without changing its operational core, while leaving quantitative characterization of the implemented online loop as a concrete next step for production-feedback integration.

## 10. Reproducibility Pointers

The full implementation underlying this paper is developed in an internal Cisco repository, while this public repository packages the manuscript, benchmark CSVs, derived reports, the public-slice dataset subtree, and a small set of historical analysis/export scripts. Public repository URL: [https://github.com/Philipzhong1980/cross-domains-evaluation-framework](https://github.com/Philipzhong1980/cross-domains-evaluation-framework). To respect privacy constraints, the packaged `meeting-summary/dataset/` subtree includes the

city_council and whitehouse_press_briefings raw artifacts but intentionally excludes raw private_data meeting contents. For the private slice, the released materials retain benchmark-level aggregates together with pseudonymous evaluation rows, but not raw meeting contents. The same boundary applies to the later focused rerun bundle, which is released only in aggregate-only form and does not redistribute meeting-level private identifiers.

Unless otherwise noted, manuscript tables are computed from meeting_model detail rows under the aggregation rule stated in Section 5.4 rather than from the convenience summary rows embedded at the top of some packaged CSVs. The public package is therefore artifact-auditable for the public slices and sufficient to verify the reported aggregate statistics against packaged artifacts, but it is **not** a standalone rerun package for the full pipeline. Even the included driver-style scripts import omitted internal src/ modules and prompt/config assets. Independent rerun of either the full merged benchmark or the public slices therefore requires the internal implementation or an equivalent reimplementation. The internal service modules, prompt/config YAML assets, and raw historical logs referenced conceptually in earlier sections are not redistributed here.

Teams seeking to instantiate this pipeline for a new generative-AI evaluation task should treat the published implementation files as entry points rather than as a monolith requiring fork-and-rewrite. The task-specific layers that require replacement are: (1) the GT schema and GT generation prompts, (2) the candidate generation configuration, and (3) the metric evaluators. The orchestration, persistence, aggregation, significance testing, and release-reporting layers can be reused unchanged. In practical terms, a new task instantiation primarily means authoring task-specific schemas and prompts rather than rebuilding the evaluation infrastructure.

The core public artifacts for this preprint are:

- Latest manuscript: meeting-summary-system-paper_v24.md, meeting-summary-system-paper_v24.html, meeting-summary-system-paper_v24.pdf
- Public artifact repository for this paper: Paper
- Main merged benchmark report and significance analysis: meeting_notes_model_comparison_combined_20260417.csv, meeting_notes_model_comparison_combined_significance_20260417.csv
- Focused three-model rerun bundle (aggregate-only): README.md, meeting_notes_model_comparison_overall_by_model_20260429_154115.csv, meeting_notes_model_comparison_overall_by_model_dataset_type_20260429_154115.csv, meeting_notes_model_comparison_summary_20260429_154115_sanitized.md
- Historical analysis/export scripts included for inspection (not standalone rerun entry points): batch_meeting_notes_model_comparison.py, deepeval_typed_empirical_comparison.py, deepeval_city_council_empirical_comparison.py, export_measured_telemetry_from_logs.py
- Typed DeepEval baseline report: deepeval_typed_empirical_comparison_all_20260421.csv
- GT agreement statistics: gt_agreement_stats_all_20260421.csv
- Single-vs-two-judge ablation: single_vs_two_judge_ablation_all_20260421.csv
- Measured telemetry from retained logs: measured_stage_level_telemetry_from_logs_20260421_131115.csv
- Architecture figures: dual_loop_quality_framework.svg, repository_pipeline_implementation.svg
- Packaged dataset subtree: dataset

## References

Entries 1-10 are scholarly references. Entries 11-13 are software/documentation sources cited for the architectural capability comparison in Table 0.

1. Philip Zhong, Kent Chen, and Don Wang. 2025. *Evaluating Embedding Models and Pipeline Optimization for AI Search Quality*. arXiv preprint arXiv:2511.22240. https://doi.org/10.48550/arXiv.2511.22240.

2. Adam Janin, Don Baron, Jane Edwards, Dan Ellis, David Gelbart, Nelson Morgan, Barbara Peskin, Thilo Pfau, Elizabeth Shriberg, Andreas Stolcke, and Chuck Wooters. 2003. The ICSI Meeting Corpus. In *Proceedings of the 2003 IEEE International Conference on Acoustics, Speech, and Signal Processing (ICASSP 2003)*, volume 1, pages 364-367. https://doi.org/10.1109/ICASSP.2003.1198793.
3. Iain McCowan, Jean Carletta, Wessel Kraaij, S. Ashby, Sandrine Bourban, Mike Flynn, Mathieu Guillemot, Thomas Hain, Jan Kadlec, Vasilis Karaiskos, Michael Kronenthal, Guillaume Lathoud, Mike Lincoln, Agnieszka Lisowska, Will Post, Dennis Reidsma, and Pete Wellner. 2005. The AMI Meeting Corpus. In *Proceedings of the 5th International Conference on Methods and Techniques in Behavioral Research*, pages 137-140. Noldus Information Technology.
4. Ming Zhong, Da Yin, Tao Yu, Ahmed Hassan Awadallah, Xipeng Qiu, and Jiawei Han. 2021. QMSum: A New Benchmark for Query-Based Multi-Domain Meeting Summarization. In *Proceedings of the 2021 Conference of the North American Chapter of the Association for Computational Linguistics: Human Language Technologies*. https://aclanthology.org/2021.naacl-main.472/.
5. Yebowen Hu, Timothy Ganter, Hanieh Deilamsalehy, Franck Dernoncourt, Hassan Foroosh, and Fei Liu. 2023. MeetingBank: A Benchmark Dataset for Meeting Summarization. In *Proceedings of the 61st Annual Meeting of the Association for Computational Linguistics (Volume 1: Long Papers)*. https://doi.org/10.18653/v1/2023.acl-long.906.
6. Soomin Kim, Seongyun Weon, Jinhwi Kim, and Hyunjoong Ko. 2023. ExplainMeetSum: A Dataset for Explainable Meeting Summarization Aligned with Human Intent. In *Proceedings of the 61st Annual Meeting of the Association for Computational Linguistics (Volume 1: Long Papers)*. https://aclanthology.org/2023.acl-long.731/.
7. Joshua Maynez, Shashi Narayan, Bernd Bohnet, and Ryan McDonald. 2020. On Faithfulness and Factuality in Abstractive Summarization. In *Proceedings of the 58th Annual Meeting of the Association for Computational Linguistics*, pages 1906-1919. Association for Computational Linguistics. https://aclanthology.org/2020.acl-main.173/.
8. Philippe Laban, Tobias Schnabel, Paul N. Bennett, and Marti A. Hearst. 2022. SummaC: Re-Visiting NLI-Based Models for Inconsistency Detection in Summarization. *Transactions of the Association for Computational Linguistics*, 10. https://doi.org/10.1162/tacl_a_00453.
9. Douwe Kiela, Max Bartolo, Yixin Nie, Divyansh Kaushik, Atticus Geiger, Julian Michael, Niloofar Mireshghallah, Khyathi Chandu, Eric Wallace, Emily Dinan, Ashish Sabharwal, and Adina Williams. 2021. Dynabench: Rethinking Benchmarking in NLP. In *Proceedings of the 2021 Conference of the North American Chapter of the Association for Computational Linguistics: Human Language Technologies*. https://aclanthology.org/2021.naacl-main.324/.
10. Shahul Es, Jithin James, Luis Espinosa-Anke, and Steven Schockaert. 2024. RAGAS: Automated Evaluation of Retrieval Augmented Generation. In *Proceedings of the 18th Conference of the European Chapter of the Association for Computational Linguistics: System Demonstrations*, pages 150-158. https://aclanthology.org/2024.eacl-demo.16/.
11. RAGAS Documentation. 2026. *Metrics Overview*. https://docs.ragas.io/en/v0.4.2/concepts/metrics/overview/. Wayback index: https://web.archive.org/web/*/https://docs.ragas.io/en/v0.4.2/concepts/metrics/overview/. Accessed April 24, 2026.
12. TruLens Documentation. 2026. *Core Concepts*. https://www.trulens.org/getting_started/core_concepts/. Wayback index: https://web.archive.org/web/*/https://www.trulens.org/getting_started/core_concepts/. Accessed April 24, 2026.
13. Confident AI Documentation. 2026. *Introduction*. https://www.confident-ai.com/docs/api-reference/introduction. Wayback index: https://web.archive.org/web/*/https://www.confident-ai.com/docs/api-reference/introduction/. Accessed April 24, 2026.